\documentclass[10pt,twocolumn,letterpaper]{article}

\usepackage{cvpr}              

\usepackage{graphicx}
\usepackage{amsmath}
\usepackage{amssymb}
\usepackage{booktabs}
\usepackage{multirow}
\usepackage[inline]{enumitem}
\usepackage{pifont}%
\newcommand{\cmark}{\ding{51}}%
\usepackage[pagebackref,breaklinks,colorlinks]{hyperref}

\usepackage[capitalize]{cleveref}
\crefname{section}{Sec.}{Secs.}
\Crefname{section}{Section}{Sections}
\Crefname{table}{Table}{Tables}
\crefname{table}{Tab.}{Tabs.}

\def\cvprPaperID{1995} 
\def\confName{CVPR}
\def\confYear{2022}

\begin{document}

\title{Beyond 3D Siamese Tracking: A Motion-Centric Paradigm for \\ 3D Single Object Tracking in Point Clouds}

\author{
    Chaoda Zheng$^{123}$,
    Xu Yan$^{123}$, 
    Haiming Zhang$^{123}$, \\ 
    Baoyuan Wang$^{4}$, 
	  Shenghui Cheng$^{5}$,  
    Shuguang Cui$^{123}$,
    Zhen Li$^{123}$\thanks{{ Corresponding author}}\\
	$^{1}$The Chinese University of Hong Kong (Shenzhen)~
  $^{2}$The future network of intelligence institute (FNII)\\
  $^{3}$Shenzhen Research Institute of Big Data~~
	$^{4}$Xiaobing.AI~~
    $^{5}$Westlake University\\
	{\tt\small	\{{chaodazheng@link.},{xuyan1@link.}, {haimingzhang@link}., {lizhen@}\}cuhk.edu.cn}
}
\maketitle

\begin{abstract}

  3D single object tracking (3D SOT) in LiDAR point clouds plays a crucial role in autonomous driving.
  Current approaches all follow the Siamese paradigm based on appearance matching.
  However, LiDAR point clouds are usually textureless and incomplete, which hinders effective appearance matching.
  Besides, previous methods greatly overlook the critical motion clues among targets.
  In this work, beyond 3D Siamese tracking, we introduce a {\textbf{motion-centric paradigm}} to handle 3D SOT from a new perspective.
  Following this paradigm, we propose a matching-free two-stage tracker {\textbf{M$^2$-Track}}.
  At the $1^{st}$-stage, $M^2$-Track localizes the target within successive frames via \textbf{m}otion transformation.
  Then it refines the target box through \textbf{m}otion-assisted shape completion at the $2^{nd}$-stage.
  Extensive experiments confirm that $M^2$-Track significantly outperforms previous state-of-the-arts on three large-scale datasets while running at \textbf{57FPS} ({\textbf{$\sim$ 8\%}}, {\textbf{$\sim$ 17\%}} and {\textbf{$\sim$ 22\%}} precision gains on KITTI, NuScenes, and Waymo Open Dataset respectively).
  Further analysis verifies each component's effectiveness and shows the motion-centric paradigm's promising potential when combined with appearance matching.
  Code will be made available at \url{https://github.com/Ghostish/Open3DSOT}.
  
\end{abstract}

\section{Introduction} 
\label{sec:intro}

Single Object Tracking (SOT) is a basic computer vision problem with various applications, such as autonomous driving~\cite{qi2021offboard,yin2020center} and surveillance system~\cite{thys2019fooling}. Its goal is to keep track of a specific target across a video sequence, given only its initial state (appearance and location).

\begin{figure}[t]
  \centering
   \includegraphics[width=1\linewidth]{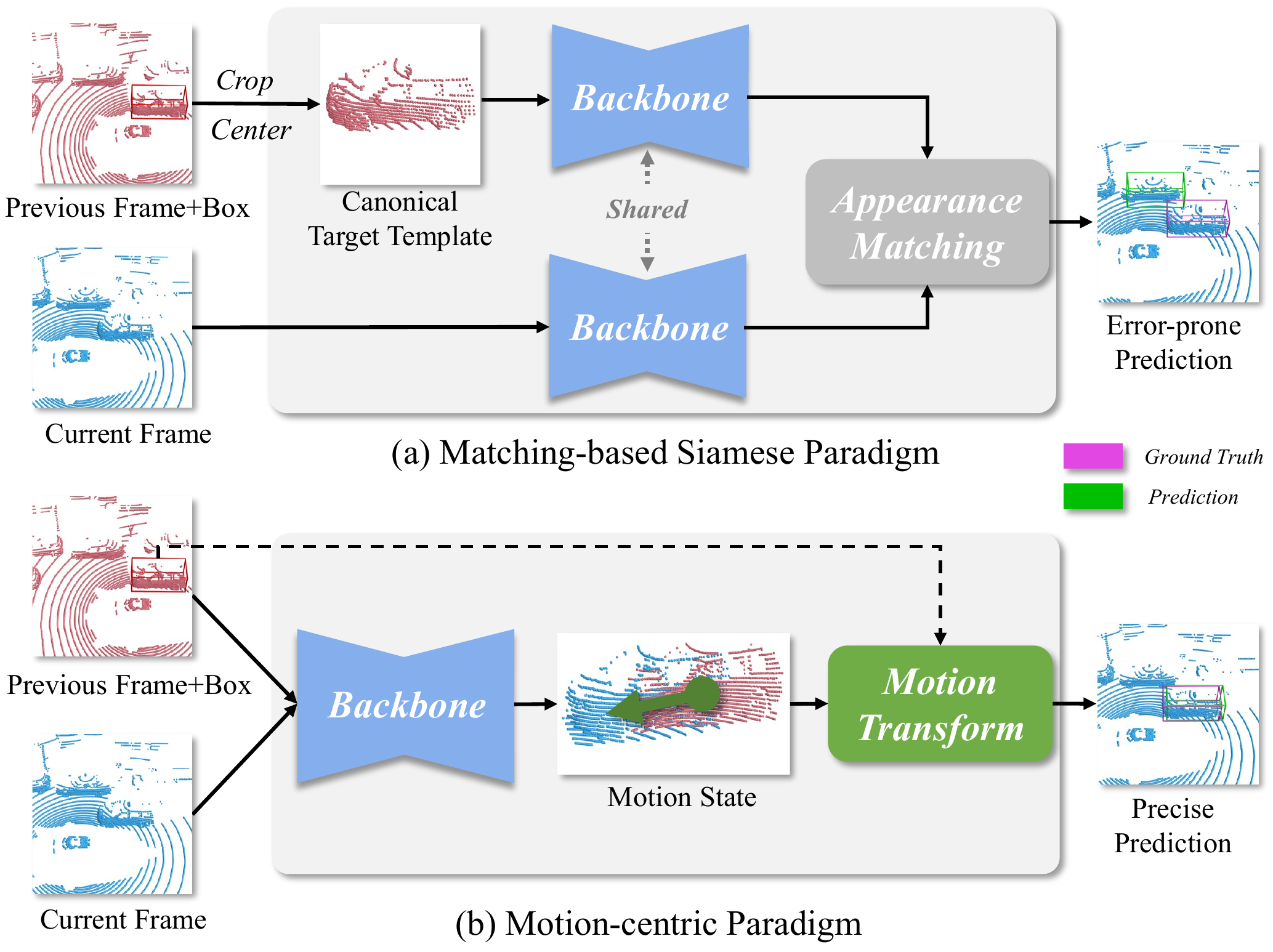}

   \caption{
     \textbf{Top}. Previous Siamese approaches obtain a canonical target template using the previous target box and search for the target in the current frame according to the matching similarity, which is sensitive to distractors. 
     \textbf{Bottom}. Our motion-centric paradigm learns the relative target motion from two consecutive frames and then robustly localizes the target in the current frame via motion transformation.}
   \label{fig:fig1}
\end{figure}

Existing LiDAR-based SOT methods~\cite{Giancola_2019_CVPR,qi2020p2b,zheng2021box,zarzar2019efficient,shan2021ptt,fang20203d} all follow the Siamese paradigm, which has been widely adopted in 2D SOT since it strikes a balance between performance and speed.
During the tracking, a Siamese model searches for the target in the candidate region with an \textbf{appearance matching} technique, which relies on the features of the target template and the search area extracted by a shared backbone (see Fig.\ref{fig:fig1}(a)).

Though the appearance matching for 3D SOT shows satisfactory results on KITTI dataset~\cite{Geiger2012CVPR}, we observe that KITTI has the following proprieties: 
\begin{enumerate*}[label=\roman*)]
    \item the target's motion between two consecutive frames is minor, which ensures no drastic appearance change;
    \item there are few/no distractors in the surrounding of the target.
\end{enumerate*}
However, the above characteristics do not hold in natural scenes. 
Due to self-occlusion, significant appearance changes may occur in consecutive LiDAR views when objects move fast, or the hardware only supports a low frame sampling rate.
 Besides, the negative samples grow significantly in dense traffic scenes. In these scenarios, it is not easy to locate a target based on its appearance alone (even for human beings).

Is the appearance matching the only solution for LiDAR SOT?
Actually, {\textbf{Motion Matters}}.
Since the task deals with a dynamic scene across a video sequence, the target's movements among successive frames are critical for effective tracking. 
Knowing this, researchers have proposed various 2D Trackers to temporally aggregate information from previous frames~\cite{wang2021transformer,bhat2020know}. However, the motion information is rarely explicitly modeled since it is hard to be estimated under the perspective distortion.
Fortunately, 3D scenes keep intact information about the object motion, which can be easily inferred from the relationships among annotated 3D bounding boxes (BBoxes)\footnote{This is greatly held for rigid objects (\eg cars), and it is approximately true for non-rigid objects (\eg pedestrian).}.
Although 3D motion matters for tracking, previous approaches have greatly overlooked it.
 Due to the Siamese paradigm, previous methods have to transform the target template (initialized by the object point cloud in the first target 3D BBox and updated with the last prediction) from the world coordinate system to its own object coordinate system. 
 This transformation ensures that the shared backbone extracts a canonical target feature, but it adversely breaks the motion connection between consecutive frames.

\begin{figure*}[t]
    \centering
     \includegraphics[width=1\linewidth]{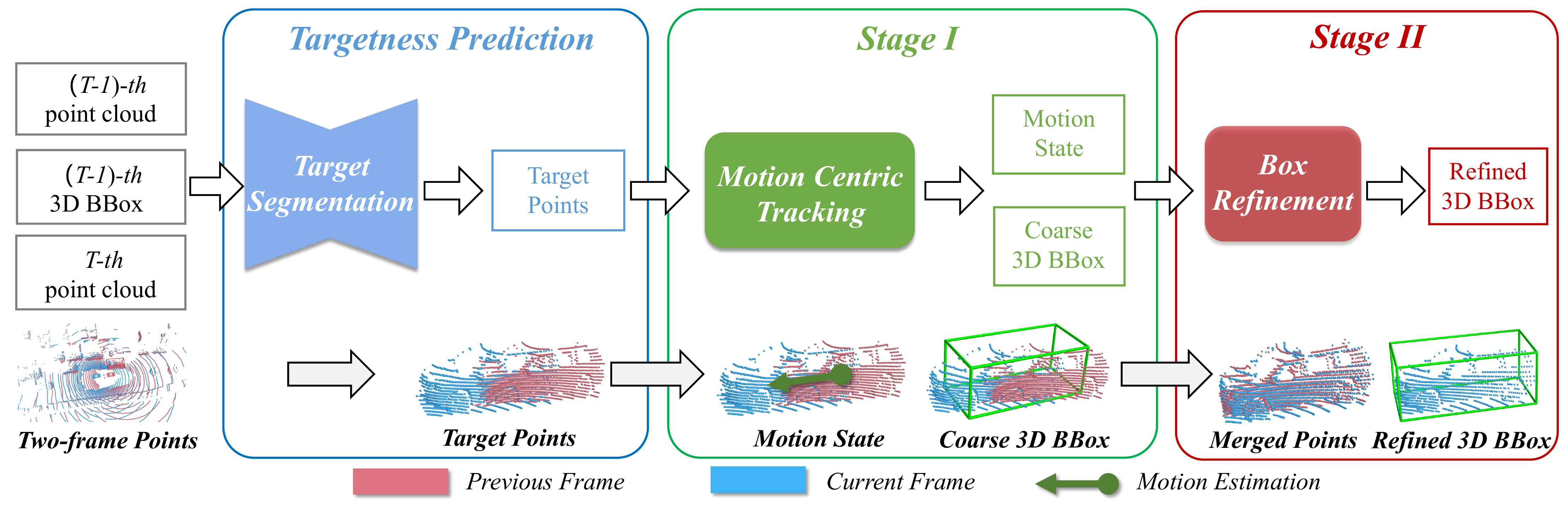}
  
     \caption{
       \textbf{The overall architecture of $M^2$-Track.}
       Given two consecutive point clouds and the possible target BBox at the previous frame, $M^2$-Track first segments the target points from their surroundings via joint spatial-temporal learning. 
       At the $1^{st}$ stage, the model takes in the target points and obtains a coarse BBox at the current frame via motion prediction and transformation.
       The coarse BBox is further refined at the $2^{nd}$ stage using motion-assisted shape completion. A detailed illustration with data flows is presented in the supplementary. 
       }
     \label{fig:pipeline}
  \end{figure*}

Based on the above observations, we propose to tackle 3D SOT from a different perspective instead of sticking to the Siamese paradigm.
For the first time, we introduce a new \textbf{motion-centric paradigm} that localizes the target in sequential frames without appearance matching by explicitly modeling the target motion between successive frames (Fig.~\ref{fig:fig1}(b)). 
Following this paradigm, we design a novel two-stage tracker {\textbf{M$^2$-Track}} (Fig.~\ref{fig:pipeline}).
During the tracking, the $1^{st}$-stage aims at generating the target BBox by predicting the inter-frame relative target motion. 
Utilizing all the information from the $1^{st}$-stage, the $2^{nd}$-stage refines the BBox using a denser target point cloud, which is aggregated from two partial target views using their relative motion.
We evaluate our model on KITTI~\cite{Geiger2012CVPR}, NuScenes\cite{caesar2020nuscenes} and Waymo Open Dataset (WOD)~\cite{sun2020scalability}, where NuScenes and WOD cover a wide variety of real-world environments and are challenging for their dense traffics. 
The experiment results demonstrate that our model outperforms the existing methods by a large margin while running faster than the previous top-performer~\cite{zheng2021box}. 
Besides, the performance gap becomes even more significant when more distractors exist in the scenes. 
Furthermore, we demonstrate that our method can directly benefit from appearance matching when integrated with existing methods.

In summary, our main contributions are as follows:
\begin{itemize}
\setlength{\itemsep}{0pt}
\setlength{\parsep}{0pt}
\setlength{\parskip}{0pt}
    \item A novel motion-centric paradigm for real-time LiDAR SOT, which is free of appearance matching.
    \item A specific second-stage pipeline named {{M$^2$-Track}} that leverages the motion-modeling and motion-assisted shape completion.
    \item State-of-the-art online tracking performance with significant improvement on three widely adopted datasets (\ie KITTI, NuScenes and Waymo Open Dataset).

  \end{itemize}
%

\section{Related Work}
\label{sec:related_work}

\noindent\textbf{Single Object Tracking.}
A majority of approaches are built for camera systems and take 2D RGB images as input~\cite{xu2020siamfc++,li2019siamrpn++,wang2021transformer,zhu2018distractor,bhat2019learning,li2018high}. Although achieving promising results, they face great challenges when dealing with low light conditions or textureless objects.
In contrast, LiDARs are insensitive to texture and robust to light variations, making them a suitable complement to cameras. 
This inspires a new trend of SOT approaches~\cite{Giancola_2019_CVPR,qi2020p2b,zheng2021box,zarzar2019efficient,shan2021ptt,fang20203d} which operate on 3D LiDAR point clouds.
These 3D methods all inherit the Siamese paradigm based on appearance matching.
As a pioneer, \cite{Giancola_2019_CVPR} uses the Kalman filter to heuristically sample a bunch of target proposals, which are then compared with the target template based on their feature similarities. The proposal which has the highest similarity with the target template is selected as the tracking result.
Since heuristic sampling is time-consuming and inhibits end-to-end training, \cite{zarzar2019efficient,qi2020p2b} propose to use a Region Proposal Network (RPN) to generate high-quality target proposals efficiently.
Unlike \cite{zarzar2019efficient} which uses an off-the-shelf 2D RPN operating on bird’s eye view (BEV), \cite{qi2020p2b} adapts SiamRPN~\cite{li2018high} to 3D point clouds by integrating a point-wise correlation operator with a point-based RPN~\cite{qi2019deep}.
The promising improvement brought by \cite{qi2020p2b} inspires a series of follow-up works~\cite{zheng2021box,shan2021ptt,fang20203d,hui20213d}. 
They focus on either improving the point-wise correlation operator~\cite{zheng2021box} by feature enhancement,
or refining the point-based RPN~\cite{fang20203d,shan2021ptt,hui20213d} with more sophisticated structures.

\noindent The appearance matching achieves excellent success in 2D SOT because images provide rich texture, which helps the model distinguish the target from its surrounding.
However, LiDAR point clouds only contain geometric appearances that lack texture information. Besides, objects in LiDAR sweeps are usually sparse and incomplete. These bring considerable ambiguities which hinder effective appearance matching.
Unlike existing 3D approaches, our work no more uses any appearance matching. Instead, we examine a new motion-centric paradigm and show its great potential for 3D SOT.

\noindent\textbf{3D Multi-object Tracking / Detection.}
In parallel with 3D SOT, 3D multi-object tracking (MOT) focuses on tracking multiple objects simultaneously. Unlike SOT where the user can specify a target of interest, MOT relies on an independent detector~\cite{shi2019pointrcnn,qi2019deep,yan2018second} to extract potential targets, which obstructs its application for unfamiliar objects (categories unknown by the detector).
Current 3D MOT methods predominantly follow the ``tracking-by-detection" paradigm, which first detects objects at each frame and then heuristically associates detected BBoxes based on objects' motion or appearance~\cite{yin2021center,Weng2020_AB3DMOT,chiu2020probabilistic,liang2020pnpnet}. Recently, \cite{luo2021exploring} proposes to jointly perform detection and tracking by combining object detection and motion association into a unified pipeline.
Our motion-centric tracker draws inspiration from the motion-based association in MOT. But unlike MOT, which applies motion estimation on detection results, our approach does not depend on any detector and can leverage the motion prediction to refine the target BBox further.

\noindent\textbf{Spatial-temporal Learning on Point Clouds.}
Our method utilizes spatial-temporal learning to infer relative motion from multiple frames.
Inspired by recent advances in natural language processing~\cite{sundermeyer2012lstm,chung2014empirical,vaswani2017attention}, there emerges methods that adapt LSTM~\cite{huang2020lstm}, GRU~\cite{yin2020lidar}, or Transformer~\cite{fan2021point} to model point cloud videos.
However, their heavy structures make them impractical to be integrated with other downstream tasks, especially for real-time applications.
Another trend forms a spatial-temporal (ST) point cloud by merging multiple point clouds with a temporal channel added to each point~\cite{rempe2020caspr,qi2021offboard,hu2020you}. Treating the temporal channel as an additional feature (like RGB or reflectance), one can process such an ST point cloud using any 3D backbones~\cite{qi2017pointnet++,qi2017pointnet} without structural modifications. We adopt this strategy to process successive frames for simplicity and efficiency.
\section{Methodology}
\label{sec:method}
\subsection{Problem Statement}\label{sec:problem}
Given the initial state of a target, our goal is to localize the target in each frame of an input sequence of a dynamic 3D scene. 
The frame at timestamp $t$ is a LiDAR point clouds ${\mathcal{P}_t \in \mathbb{R}^{N_t \times 3}}$ with $N_t$ points and $3$ channels, where the point channels encode the $xyz$ global coordinate.
The initial state of the target is given as its 3D BBox at the first frame $\mathcal{P}_1$. A 3D BBox $\mathcal{B}_t \in \mathbb{R}^7$ is parameterized by its center ($xyz$ coordinate), orientation (heading angle $\theta$ around the $up$-axis), and size (width, length, and height).  For the tracking task, we further assume that the size of a target remains unchanged across frames even for non-rigid objects (for a non-rigid object, its BBox size is defined by its maximum extent in the scene).
For each frame $\mathcal{P}_t$, the tracker outputs the amodal 3D BBox of the target with access to only history frames $\{\mathcal{P}_i\}_{i=1}^t$. 

\subsection{Motion-centric Paradigm}\label{sec:paradigm}
Given an input LiDAR sequence and the 3D BBox of the target in the first frame, the motion-centric tracker aims to localize the target frame by frame using explicit motion modeling. 
At timestamp $t$ ($t > 1$), the target BBox $\mathcal{B}_{t-1}$ at frame $t-1$ is known (either given as the initial state or predicted by the tracker).
Having two consecutive frame $\mathcal{P}_t$ and $\mathcal{P}_{t-1}$ as well as the target BBox $\mathcal{B}_{t-1}$ in $\mathcal{P}_{t-1}$, the tracker predicts the \textbf{relative target motion (RTM)} between the successive two frames. We only consider 4DOF instead of 6DOF RTM since the targets are always aligned with the ground plane (no roll and pitch). Specifically, a 4DOF RTM $\mathcal{M}_{t-1,t} \in \mathbb{R}^4$ is defined between two target BBoxes in frame $t$ and $t-1$, and contains the translation offset $(\Delta x,\Delta y,\Delta z)$ and the yaw offset $\Delta \theta$.
We can formulate this process as a function $\mathcal{F}$:
\begin{equation}
\begin{aligned}
\mathcal{F}&: \mathbb{R}^{N_t \times C} \times \mathbb{R}^{N_{t-1} \times C} \times \mathbb{R}^7\mapsto \mathbb{R}^4, \\
\mathcal{F}&(\mathcal{P}_t,\mathcal{P}_{t-1},\mathcal{B}_{t-1}) \mapsto  (\Delta x,\Delta y,\Delta z,\Delta \theta);
\end{aligned}
\end{equation}
Having the predicted RTM $\mathcal{M}_{t-1,t}$, one can easily obtain the target BBox in $\mathcal{P}_t$ using rigid body transformation:
\begin{equation}
  \mathcal{B}_{t} = {\text{Transform}}(\mathcal{B}_{t-1},\mathcal{M}_{t-1,t}).
  \label{eq:rigid_body_transform}
\end{equation}
\subsection{$M^2$-Track: Motion-Centric Tracking Pipeline}\label{sec:pipeline}
Following the motion-centric paradigm, we design a two-stage motion-centric tracking pipeline $M^2$-Track (illustrated in Fig.\ref{fig:pipeline}).
$M^2$-Track first coarsely localizes the target through target segmentation and \textbf{m}otion transformation at the $1^{st}$ stage, and then refines the BBox at the $2^{nd}$ stage using \textbf{m}otion-assisted shape completion.
More details of each module are given as follows.

\noindent\textbf{Target segmentation with spatial-temporal learning} 

\noindent
To learn the relative target motion, we first need to segment the target points from their surrounding. 
Taking as inputs two consecutive frames $\mathcal{P}_{t}$ and $\mathcal{P}_{t-1}$ together with the target BBox $\mathcal{B}_{t-1}$,  we do this by exploiting the spatial-temporal relation between the two frames (illustrated in the first part of Fig.\ref{fig:pipeline}).
Similar to \cite{rempe2020caspr,piergiovanni20214d}, we construct a spatial-temporal point cloud $\mathcal{P}_{t-1,t} \in \mathbb{R}^{(N_{t-1} + N_{t}) \times 4} = \{p_i = (x_i,y_i,z_i,t_i)\}_{i=1}^{N_{t-1}+N_t}$ from $\mathcal{P}_{t-1}$ and $\mathcal{P}_{t}$ by adding a temporal channel to each point and then merging them together.
Since there are multiple objects in a scene, we have to specify the target of interest according to $\mathcal{B}_{t-1}$.
To this end, we create a prior-targetness map $\mathcal{S}_{t-1,t} \in \mathbb{R}^{N_{t-1} + N_t}$ to indicate target location in $\mathcal{P}_{t-1,t}$, where $s_i \in \mathcal{S}_{t-1,t}$ is defined as:
\begin{equation}
  \begin{aligned}
  s_i = 
  \left\{
    \begin{matrix}
      0& \text{ if } p_i \text{ is in } \mathcal{P}_{t-1} \text{ and } p_i \text{ is not in }\mathcal{B}_{t-1}\\
      1& \text{ if } p_i \text{ is in } \mathcal{P}_{t-1} \text{ and } p_i \text{ is in }\mathcal{B}_{t-1}\\
      0.5& \text{ if } p_i \text{ is in } \mathcal{P}_{t}
    \end{matrix}\right.
  \end{aligned}
\end{equation}
Intuitively, one can regard $s_i$ as the prior-confidence of $p_i$ being a target point. For a point in $\mathcal{P}_{t-1}$, we set its confidence according to its location with respect to $\mathcal{B}_{t-1}$. Since the target state in $\mathcal{P}_{t}$ is unknown, we set a median score 0.5 for each point in $\mathcal{P}_{t}$.
Note that $\mathcal{S}_{t-1,t}$ is not 100\% correct for points in $\mathcal{P}_{t-1}$ since $\mathcal{B}_{t-1}$ could be the previous output by the tracker.
After that, we form a 5D point cloud by concatenating $\mathcal{P}_{t-1,t}$ and $\mathcal{S}_{t-1,t}$ along the channel axis, and use a PointNet~\cite{qi2017pointnet} segmentation network to obtain the target mask, which is finally used to extract a spatial-temporal target point cloud $\mathcal{\widetilde{P}}_{t-1,t} \in \mathbb{R}^{({M_{t-1} + M_{t})} \times 4}$, where $M_{t-1}$ and $M_{t}$ are the numbers of target points in frame ($t-1$) and $t$ respectively.

 \begin{figure}[t]
   \centering
    \includegraphics[width=0.95\linewidth]{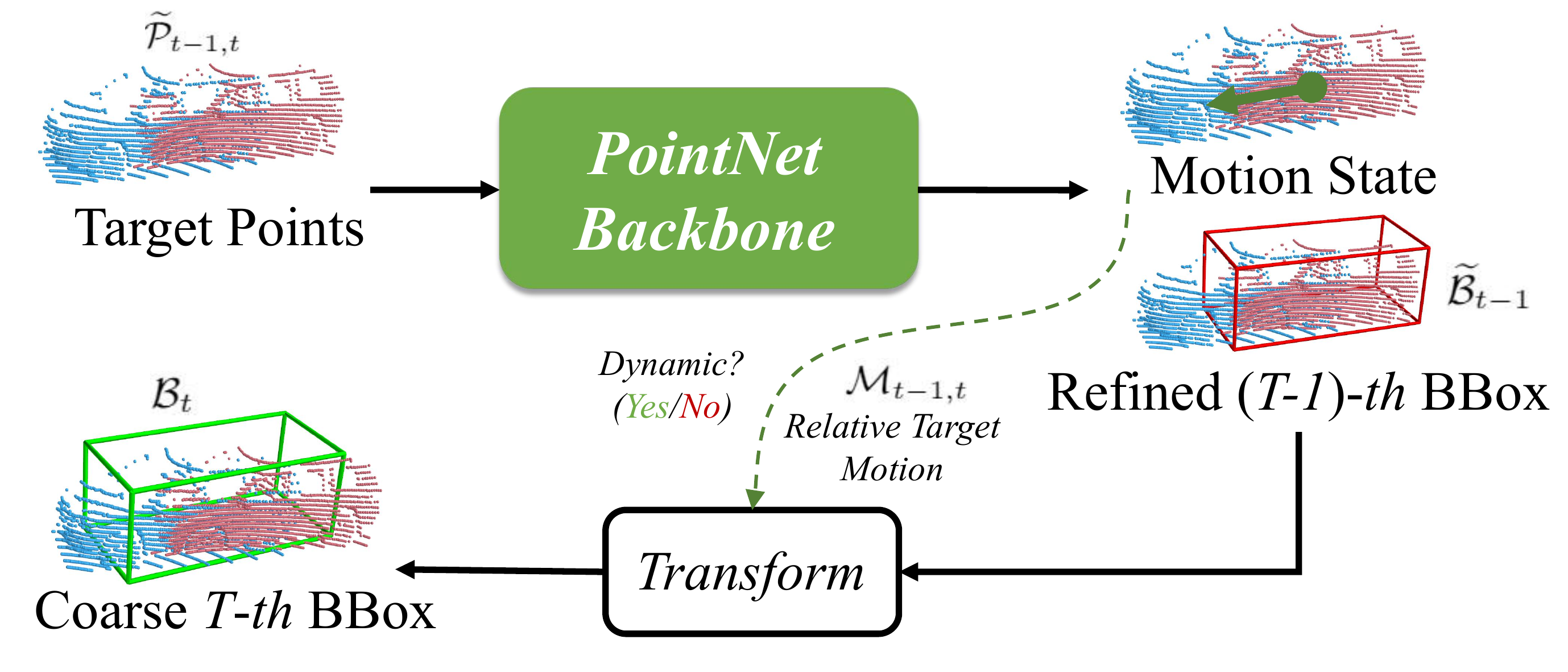}
 
    \caption{
      \textbf{Stage I.}
      Taking in the segmented target points $\mathcal{\widetilde{P}}_{t-1,t}$ and the target BBox $\mathcal{{B}}_{t-1}$ at the previous frame, the model outputs a relative target motion state ( including a RTM $\mathcal{M}_{t-1,t}$ and 2D binary motion state logits), a refined target BBox $\mathcal{\widetilde{B}}_{t-1}$ at the previous frame, and a coarse target BBox $\mathcal{B}_{t}$ at the current frame.
      }
    \label{fig:stage1}
 \end{figure}
 \begin{figure}[t]
  \centering
   \includegraphics[width=0.95\linewidth]{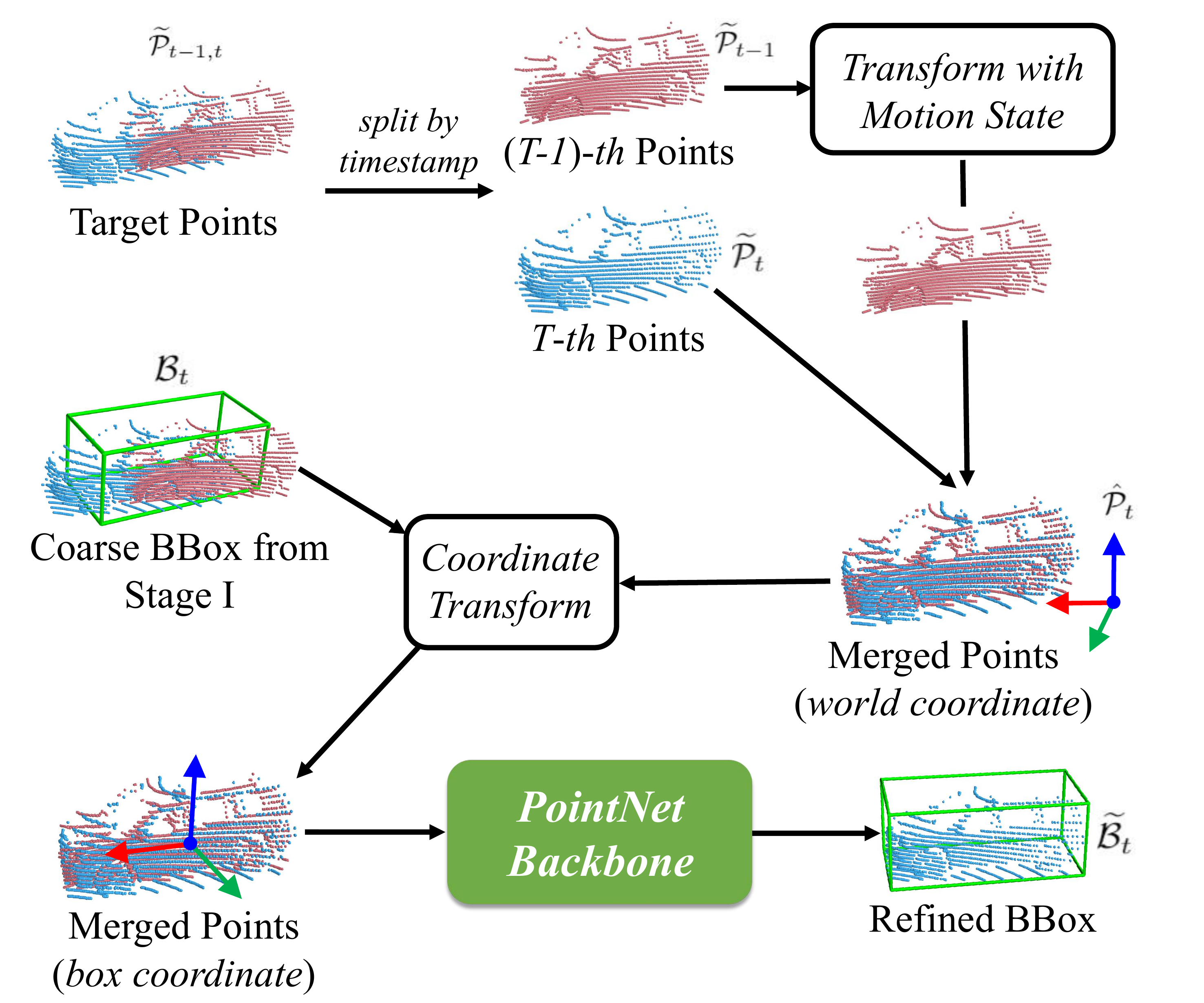}

   \caption{
    \textbf{Stage II.}
    Taking the segmented target points $\mathcal{\widetilde{P}}_{t-1,t}$ and the coarse target BBox $\mathcal{B}_{t}$ as inputs, the model regresses a refined target BBox $\mathcal{\widetilde{B}}_{t}$ on a denser point cloud, which is merged from two partial target point clouds according to their relative motion state.
     }
   \label{fig:stage2}
\end{figure}

\noindent\textbf{Stage I: Motion-Centric BBox prediction}\\
\noindent As shown in Fig.~\ref{fig:stage1}, we encode the spatial-temporal target point clouds $\mathcal{\widetilde{P}}_{t-1,t}$ into an embedding using another PointNet encoder. A multi-layer perceptron (MLP) is applied on top of the embedding to obtain the motion state of the target, which includes a 4D RTM $\mathcal{M}_{t-1,t}$ and 2D binary classification logits indicating whether the target is dynamic. To reduce accumulation errors while performing frame-by-frame tracking, we generate a refined previous target BBox $\mathcal{\widetilde{B}}_{t-1}$ by predicting its RTM with respect to $\mathcal{B}_{t-1}$ through another MLP (More details are presented in the supplementary). Finally, we can get the current target BBox $\mathcal{B}_{t}$ by applying Eqn.~\ref{eq:rigid_body_transform} on $\mathcal{M}_{t-1,t}$ and $\mathcal{\widetilde{B}}_{t-1}$ if the target is classified as dynamic. Otherwise, we simply set $\mathcal{B}_{t}$ as $\mathcal{\widetilde{B}}_{t-1}$.

\noindent\textbf{Stage II: BBox refinement with shape completion}\\
\noindent Inspired by two-stage detection networks~\cite{shi2019pointrcnn,shi2020points}, we improve the quality of the $1^{st}$-stage BBox $\mathcal{B}_{t}$ by additionally regressing a relative offset, which can be regarded as a RTM between $\mathcal{B}_{t}$ and the refined BBox $\mathcal{\widetilde{B}}_{t}$.
Unlike detection networks, we refine the BBox via a novel \textbf{motion-assisted shape completion} strategy.
Due to self-occlusion and sensor movements, LiDAR point clouds suffer from great incompleteness, which hinders precise BBox regression. To mitigate this, we form a denser target point cloud by using the predicted motion state to aggregate the targets from two successive frames.
According to the temporal channel, two target point clouds $\mathcal{\widetilde{P}}_{t-1} \in \mathbb{R}^{M_{t-1} \times 3}$ and $\mathcal{\widetilde{P}}_{t} \in \mathbb{R}^{M_t \times 3}$ from different timestamps are extracted from $\mathcal{\widetilde{P}}_{t-1,t}$.
Based on the motion state, we transform $\mathcal{\widetilde{P}}_{t-1}$ to the current timestamp using $\mathcal{M}_{t-1,t}$ if the target is dynamic. The transformed point cloud (identical as $\mathcal{\widetilde{P}}_{t-1}$ if the target is static) is merged with  $\mathcal{\widetilde{P}}_{t}$ to form a denser point cloud $\mathcal{\hat{P}}_{t} \in \mathbb{R}^{(M_{t-1} + M_t) \times 3}$. Similar to \cite{shi2019pointrcnn,qi2021offboard}, we transform $\mathcal{\hat{P}}_{t}$ from the world coordinate system to the canonical coordinate system defined by $\mathcal{B}_{t}$.
We apply a PointNet on the canonical $\mathcal{\hat{P}}_{t}$ to regress another RTM with respect to $\mathcal{B}_{t}$. Finally, the refined target BBox $\mathcal{\widetilde{B}}_{t}$ is obtained by applying Eqn.~\ref{eq:rigid_body_transform}.

\subsection{Box-aware Feature Enhancement}
As shown in \cite{zheng2021box}, LiDAR SOT directly benefits from the part-aware and size-aware information, which can be depicted by point-to-box relation.
To achieve better target segmentation, we construct a distance map $\mathcal{C}_{t-1} \in \mathbb{R}^{N_{t-1} \times 9}$ by computing the pairwise Euclidean distance between $\mathcal{P}_{t-1}$ and 9 key points of $\mathcal{B}_{t-1}$ (eight corners and one center arranged in a predefined order with respect to the canonical box coordinate system). 
After that, we extend $\mathcal{C}_{t-1}$ to size $(N_{t-1} + N_t) \times 9$ with zero-padding (for points in $\mathcal{P}_{t}$) and
additionally concatenate it with $\mathcal{P}_{t-1,t}$ and $\mathcal{S}_{t-1,t}$.
The overall box-aware features are then sent to the PointNet segmentation network to obtain better target segmentation.

\subsection{Implementation Details}\label{sec:impl_detail}
\noindent\textbf{Loss Functions.}
The loss function contains classification losses and regression losses, which is defined as 
$L = \lambda_1 L_{\text{cls\_target}} + \lambda_2 L_{\text{cls\_motion}} + \lambda_3 (L_{\text{reg\_motion}} + L_{\text{reg\_refine\_prev}} + L_{\text{reg\_1st}} + L_{\text{reg\_2nd}})$.
$L_{\text{cls\_target}}$ and $L_{\text{cls\_motion}}$ are standard cross-entropy losses for target segmentation and motion state classification at the $1^{st}$-stage (Points are considered as the target if they are inside the target BBoxes; A target is regarded as dynamic if its center moves more than 0.15 meter between two frames).
All regression losses are defined as the Huber loss~\cite{ren2015faster} between the predicted and ground-truth RTMs (inferred from ground-truth target BBoxes),
where $L_{\text{reg\_motion}}$ is for the RTM between targets in the two frames; $L_{\text{reg\_refine\_prev}}$ is for the RTM between the predicted and the ground-truth BBoxes at timestamp $(t-1)$; $L_{\text{reg\_1st}}$ / $L_{\text{reg\_2nd}}$ is for the RTM between the $1^{st}$ / $2^{nd}$-stage and ground-truth BBoxes.
We empirically set $\lambda_1 = \lambda_2 = 0.1$ and $\lambda_3 = 1$.

\noindent\textbf{Input \& Motion Augmentation.}
Since SOT only takes care of one target in a scene, we only need to consider a subregion where the target may appear. For two consecutive frames at $(t-1)$ and $t$ timestamp, we choose the subregion by enlarging the target BBox at $(t-1)$ timestamp by 2 meters. We then sample 1024 points from the subregion respectively at $(t-1)$ and $t$ timestamp to form $\mathcal{P}_{t-1}$ and $\mathcal{P}_{t}$. To simulate testing errors during the training, we feed the model a perturbed BBox by adding a slight random shift to the ground-truth target BBox at $(t-1)$ timestamp.
To encourage the model to learn various motions during the training, we randomly flip both targets' points and BBoxes in their horizontal axes and rotate them around their $up$-axes by \textit{Uniform}[$-10^{\circ}$,$10^{\circ}$]. We also randomly translate the targets by offsets drawn from \textit{Uniform} [-0.3, 0.3] meter.

\noindent\textbf{Training \& Inference.}
We train our models using the Adam optimizer with batch size 256 and an initial learning rate 0.001, which is decayed by 10 times every 20 epochs. The training takes $\sim 4$ hours to converge on a V100 GPU for the KITTI Cars. During the inference, the model tracks a target frame-by-frame in a point cloud sequence given the target BBox at the first frame.
\begin{table}
  \renewcommand\tabcolsep{4pt} 
  \footnotesize
  \footnotesize
  \caption{Comparison among our $M^2$-Track and the state-of-the-art methods on the KITTI datasets. \textit{Mean} shows the average result weighed by frame numbers.
  \textbf{Bold} and \underline{underline} denote the best performance and the second-best performance, respectively. Improvements over previous state-of-the-arts are shown in \textit{Italic}.}
    \vspace{-0.3cm}
    \begin{center}
    \begin{tabular}{lc|ccccc}
        \toprule[.05cm]
        \multirow{2}*{}
        & Category & Car   & Pedestrian & Van & Cyclist &Mean \\
        & Frame Number &\textit{6424} &\textit{6088} &\textit{1248} &\textit{308} &\textit{14068} \\
        \midrule
        \midrule
        \multirow{9}*{\rotatebox{90}{Success}}
        & SC3D~\cite{Giancola_2019_CVPR} &41.3 & 18.2 &40.4 &41.5 &31.2 \\
        & SC3D-RPN~\cite{zarzar2019efficient} &36.3 & 17.9 & - &{43.2} & -  \\
        & P2B~\cite{qi2020p2b} &56.2 &28.7 &40.8 &32.1 &42.4 \\
        & 3DSiamRPN~\cite{fang20203d} &{58.2} &{35.2} &{45.6} &36.1 &{46.6} \\
        & LTTR~\cite{cui20213d} &{65.0} &{33.2} &{35.8} &\underline{66.2} &{48.7} \\
        & PTT~\cite{shan2021ptt} &\underline{67.8} &{44.9} &{43.6} &37.2 & {55.1} \\
        & V2B~\cite{hui20213d} &\textbf{70.5} &\underline{48.3} &{50.1} &40.8 &\underline{58.4}\\
        & BAT~\cite{zheng2021box} &{65.4} &{45.7} &\underline{52.4} &33.7 &{55.0} \\
        \cmidrule{2-7}

        & $M^2$-Track (Ours) &{65.5} &\textbf{61.5} &\textbf{53.8} &\textbf{73.2} &\textbf{62.9}\\
        & \textit{Improvement} &\textit{\textcolor[rgb]{0.7,0.0,0.0}{$\downarrow $5.0}} &\textit{\textcolor[rgb]{0.1,0.7,0.0}{$\uparrow$13.2}} &\textit{\textcolor[rgb]{0.1,0.7,0.0}{$\uparrow$1.4}} &\textit{\textcolor[rgb]{0.1,0.7,0.0}{$\uparrow$7.0}} &\textit{\textcolor[rgb]{0.1,0.7,0.0}{$\uparrow$4.5}}\\
        \midrule
        \midrule
        \multirow{9}*{\rotatebox{90}{Precision}}
        & SC3D~\cite{Giancola_2019_CVPR} &57.9 & 37.8 &47.0 &70.4 &48.5 \\
        & SC3D-RPN~\cite{zarzar2019efficient} &51.0 & 47.8 &- &{81.2} &-  \\
        & P2B~\cite{qi2020p2b} &72.8 &49.6 &48.4 &44.7 &60.0 \\
        & 3DSiamRPN~\cite{fang20203d} &{76.2} &{56.2} &{52.8} &49.0 &{64.9} \\
        & LTTR~\cite{cui20213d} &{77.1} &{56.8} &{45.6} &\underline{89.9} &{65.8} \\
        & PTT~\cite{shan2021ptt} &\textbf{81.8} &{72.0} &{52.5} &47.3 &{74.2}\\
        & V2B~\cite{hui20213d} &\underline{81.3} &{73.5} &{58.0} &49.7 &\underline{75.2}\\
        & BAT~\cite{zheng2021box} &{78.9} &\underline{74.5} &\underline{67.0} &45.4 &\underline{75.2} \\
        \cmidrule{2-7}
        & $M^2$-Track (Ours) &{80.8} &\textbf{88.2} &\textbf{70.7} &\textbf{93.5} &\textbf{83.4}\\
        & \textit{Improvement} &\textit{\textcolor[rgb]{0.7,0,0}{$\downarrow$ 1.0}} &\textit{\textcolor[rgb]{0.1,0.7,0.0}{$\uparrow$13.7}} &\textit{\textcolor[rgb]{0.1,0.7,0.0}{$\uparrow$3.7}} &\textit{\textcolor[rgb]{0.1,0.7,0.0}{$\uparrow$3.6}} &\textit{\textcolor[rgb]{0.1,0.7,0.0}{$\uparrow$8.2}}\\
        \toprule[.05cm]
  
      \end{tabular}
    \end{center}
    
    \label{tab:kitti}
   		\vspace{-0.3cm}
    \end{table}
%
\section{Experiments}
\label{sec:exp}

\begin{table*}
  \renewcommand\tabcolsep{8pt} 
  \footnotesize
  \caption{Comparison of $M^2$-Track against state-of-the-arts on the NuScenes and Waymo Open Dataset using the same annotations in Tab.~\ref{tab:kitti}.} 
    \begin{center}

    \begin{tabular}{lc|cccccc|ccc}
        \toprule[.05cm]
        \multirow{3}*{}
        &  Dataset  & \multicolumn{6}{c|}{{NuScenes}} & \multicolumn{3}{c}{{Waymo Open Dataset}}\\ 
        & Category  & Car & Pedestrian & Truck & Trailer & Bus  & Mean & Vehicle   & Pedestrian  &Mean\\
        & Frame Number &\textit{64,159} & \textit{33,227} &\textit{13,587} &\textit{3,352} &\textit{2,953} &\textit{117,278} &\textit{1,057,651} &\textit{510,533} &\textit{1,568,184} \\
        \midrule
        \midrule
        \multirow{5}*{\rotatebox{90}{Success}}
        & SC3D~\cite{Giancola_2019_CVPR} & 22.31 & 11.29 & 30.67 & 35.28 & 29.35 & 20.70 &- & - &- \\
        & P2B~\cite{qi2020p2b} &38.81 & 28.39 &42.95 &48.96 &32.95 &36.48 &28.32 & 15.60 & 24.18\\
        & BAT~\cite{zheng2021box} &\underline{40.73} & \underline{28.83} &\underline{45.34} &\underline{52.59} &\underline{35.44} & \underline{38.10} &\underline{35.62} &\underline{22.05} &\underline{31.20} \\
        \cmidrule{2-11}
        & $M^2$-Track (Ours) &\textbf{55.85} & \textbf{32.10} &\textbf{57.36} &\textbf{57.61} &\textbf{51.39} & \textbf{49.23} &\textbf{43.62} &\textbf{42.10} &\textbf{43.13} \\
        & \textit{Improvement} &\textcolor[rgb]{0.1,0.7,0.0}{\textit{$\uparrow$15.12}} & \textcolor[rgb]{0.1,0.7,0.0}{\textit{$\uparrow$3.27}} &\textcolor[rgb]{0.1,0.7,0.0}{\textit{$\uparrow$12.02}} &\textcolor[rgb]{0.1,0.7,0.0}{\textit{$\uparrow$5.02}} &\textcolor[rgb]{0.1,0.7,0.0}{\textit{$\uparrow$15.95}} & \textcolor[rgb]{0.1,0.7,0.0}{\textit{$\uparrow$11.14}} &\textcolor[rgb]{0.1,0.7,0.0}{\textit{$\uparrow$8.00}} &\textcolor[rgb]{0.1,0.7,0.0}{\textit{$\uparrow$20.05}} &\textcolor[rgb]{0.1,0.7,0.0}{\textit{$\uparrow$11.92}} \\
        \midrule
        \midrule
        \multirow{5}*{\rotatebox{90}{Precision}}
        & SC3D~\cite{Giancola_2019_CVPR} & 21.93 & 12.65 & 27.73 & 28.12 & 24.08 & 20.20 &- & - &-\\
        & P2B~\cite{qi2020p2b} &43.18 & 52.24 &41.59 &40.05 &27.41 &45.08 &35.41 & 29.56 & 33.51 \\
        & BAT~\cite{zheng2021box} &\underline{43.29} & \underline{53.32} &\underline{42.58} &\underline{44.89} &\underline{28.01} &\underline{45.71} &\underline{44.15} &\underline{36.79} &\underline{41.75}\\
        \cmidrule{2-11}
        & $M^2$-Track (Ours) &\textbf{65.09} & \textbf{60.92} &\textbf{59.54} &\textbf{58.26} &\textbf{51.44} &\textbf{62.73} &\textbf{61.64} &\textbf{67.31} &\textbf{63.48}\\
        & \textit{Improvement} &\textcolor[rgb]{0.1,0.7,0.0}{\textit{$\uparrow$21.80}} & \textcolor[rgb]{0.1,0.7,0.0}{\textit{$\uparrow$7.60}} &\textcolor[rgb]{0.1,0.7,0.0}{\textit{$\uparrow$16.96}} &\textcolor[rgb]{0.1,0.7,0.0}{\textit{$\uparrow$13.37}} &\textcolor[rgb]{0.1,0.7,0.0}{\textit{$\uparrow$23.43}} & \textcolor[rgb]{0.1,0.7,0.0}{\textit{$\uparrow$17.02}} &\textcolor[rgb]{0.1,0.7,0.0}{\textit{$\uparrow$17.49}} &\textcolor[rgb]{0.1,0.7,0.0}{\textit{$\uparrow$30.52}} &\textcolor[rgb]{0.1,0.7,0.0}{\textit{$\uparrow$21.73}} \\
        \toprule[.05cm]

      \end{tabular}
    \end{center}
    
    \label{tab:nusc_waymo}
    \vspace{-0.3cm}

\end{table*}
\begin{figure*}[t]
  \centering
   \includegraphics[width=1\linewidth]{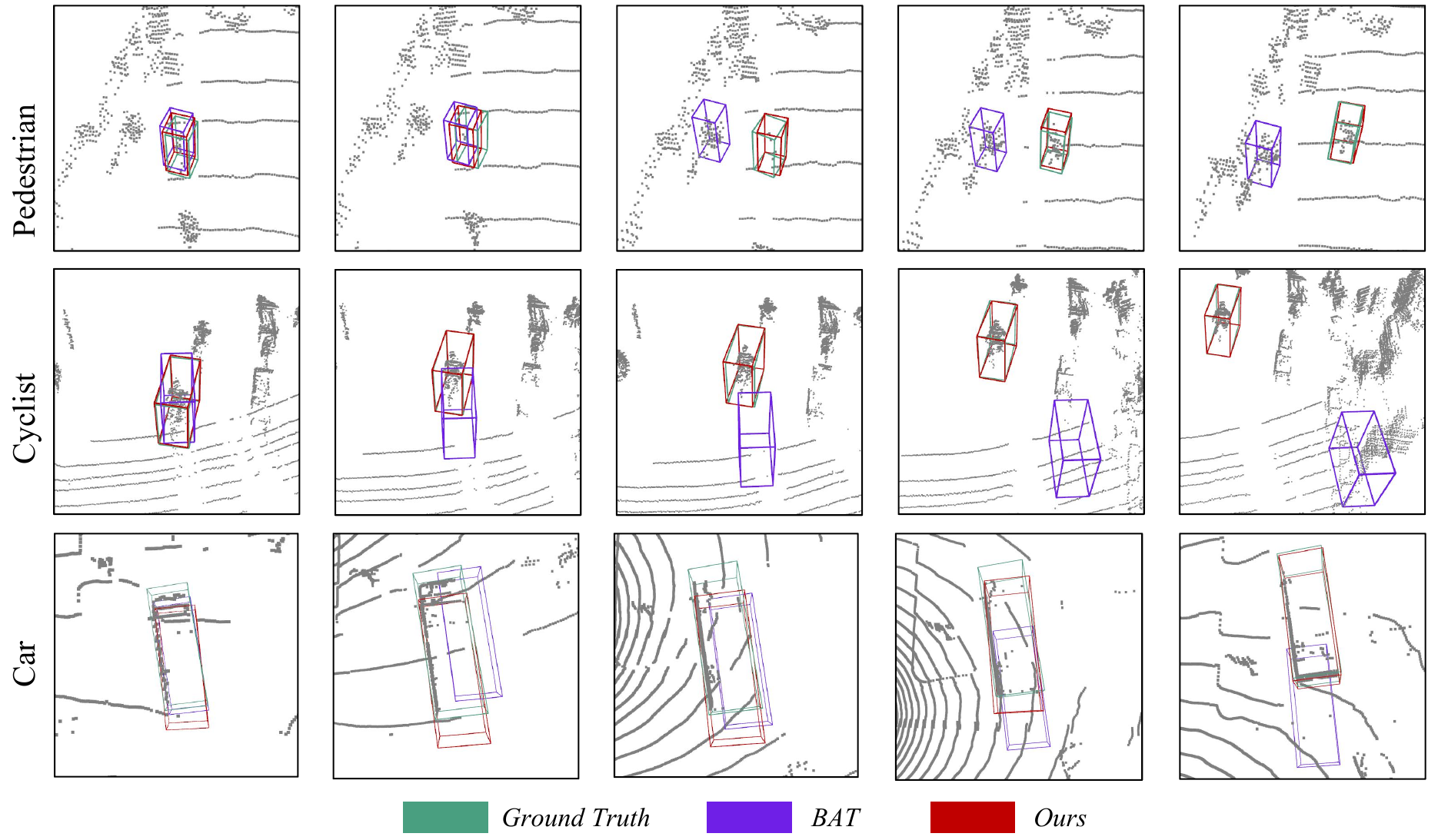}

   \caption{Visualization results. \textbf{Top:} Distractor case in KITTI. \textbf{Middle:} Large motion case in KITTI. \textbf{Bottom:} Case in NuScenes.}
   \label{fig:visualization}
\end{figure*}
\subsection{Experiment Setups}
\noindent\textbf{Datasets.} We extensively evaluate our approach on three large-scale datasets: KITTI~\cite{Geiger2012CVPR}, NuScenes\cite{caesar2020nuscenes} and Waymo Open Dataset (WOD)~\cite{sun2020scalability}. 
We follow~\cite{Giancola_2019_CVPR} to adapt these datasets for 3D SOT by extracting the tracklets of annotated tracked instances from each of the scenes.
\textbf{KITTI} contains 21 training sequences and 29 test sequences. We follow previous works~\cite{Giancola_2019_CVPR,qi2020p2b,zheng2021box} to split the training set into train/val/test splits due to the inaccessibility of the test labels.
\textbf{NuScenes} contains 1000 scenes, which are divided into 700/150/150 scenes for train/val/test. Officially, the train set is further evenly split into ``train\_track" and ``train\_detect" to remedy overfitting. Following~\cite{zheng2021box}, we train our model with ``train\_track" split and test it on the val set. 
\textbf{WOD} includes 1150 scenes with 798 for training, 202 for validation, and 150 for testing. We do training and testing respectively on the training and validation set.
Note that NuScenes and WOD are much more challenging than KITTI due to larger data volumes and complexities.
The LiDAR sequences are sampled at 10Hz for both KITTI and WOD.
Though NuScenes samples at 20Hz, it only provides the annotations at 2Hz.
Since only annotated keyframes are considered, such a lower frequency for keyframes introduces additional difficulties for NuScenes.

\noindent\textbf{Evaluation Metrics.} We evaluate the models using the One Pass Evaluation (OPE)~\cite{wu2013online}. It defines \textit{overlap} as the Intersection Over Union (IOU) between the predicted and ground-truth BBox, and defines \textit{error} as the distance between two BBox centers. We report the \textit{Success} and \textit{Precision} of each model in the following experiments. \textit{Success} is the Area Under the Curve (AUC) with the \textit{overlap} threshold varying from 0 to 1. \textit{Precision} is the AUC with the \textit{error} threshold from 0 to 2 meters.

\subsection{Comparison with State-of-the-arts}
\noindent\textbf{Results on KITTI.}
We compare $M^2$-Track with seven top-performance approaches~\cite{Giancola_2019_CVPR,zarzar2019efficient,qi2020p2b,fang20203d,cui20213d,shan2021ptt,zheng2021box,hui20213d}, which have published results on KITTI.
As shown in Tab.~\ref{tab:kitti}, our method benefits both rigid and non-rigid object tracking, outperforming current approaches under all categories except Car, where PTT~\cite{shan2021ptt} and V2B~\cite{hui20213d} surpass us by minor margins.
The lack of car distractors in the scenes makes our improvement over previous appearance-matching-based methods minor for cars.
But our improvement for pedestrians is significant (13.2\%/13.7\% in terms of success/precision) because pedestrian distractors are widespread in the scenes (see the supplementary for more details).
Besides, methods using point-based RPN~\cite{qi2020p2b,fang20203d,shan2021ptt,zheng2021box} all perform badly on cyclists, which are relative small in size but usually move fast across time. 
The second row in Fig.~\ref{fig:visualization} shows the case in which a cyclist moves rapidly across frames. Our method perfectly keeps track of the target while BAT almost fails.
To handle such fast-moving objects, \cite{zarzar2019efficient,cui20213d} leverage BEV-based RPN to generate high-recall proposals from a larger search region.
In contrast, we handle this simply by motion modeling without sophisticated architectures.
%

\begin{table}[t]
  \centering
  \footnotesize
  \caption{Influence of Motion Augmentation. ``aug" stands for motion augmentation.}
  \vspace{-0.3cm}
    \begin{tabular}{c|cc}
    \toprule
    Method & Success & Precision \\
    \midrule\midrule
    BAT~\cite{zheng2021box} w/o aug& 65.37  & 78.88  \\
    BAT~\cite{zheng2021box} w/ aug & 63.59 \textcolor[rgb]{0.7,0.0,0.0}{\small $\downarrow$ 1.78} &  76.99 \textcolor[rgb]{0.7,0.0,0.0}{\small $\downarrow$ 1.89} \\
    \midrule
    P2B~\cite{qi2020p2b} w/o aug& 56.20 & 72.80 \\
    P2B~\cite{qi2020p2b} w/ aug &  55.21 \textcolor[rgb]{0.7,0.0,0.0}{\small $\downarrow$ 0.99} &  71.51 \textcolor[rgb]{0.7,0.0,0.0}{\small $\downarrow$ 1.29} \\
    \midrule
    $M^2$-Track w/o aug & 65.29    & 77.12  \\
    $M^2$-Track w/ aug & 65.49 \textcolor[rgb]{0.1,0.7,0.0}{\small $\uparrow$ 0.20} & 80.81 \textcolor[rgb]{0.1,0.7,0.0}{\small $\uparrow$ 3.69}\\
    \bottomrule
    \end{tabular}%
  \label{tab:augmentation}%
  \vspace{-0.3cm}
\end{table}%

\noindent\textbf{Results on NuScenes \& WOD.}
We select three representative open-source works: SC3D~\cite{Giancola_2019_CVPR}, P2B~\cite{qi2020p2b} and BAT~\cite{zheng2021box} as our competitors on NuScenes and WOD. The results on NuScenes except for the Pedestrian class are provided by~\cite{zheng2021box}. We use the published codes of the competitors to obtain other results absent in~\cite{zheng2021box}.
SC3D~\cite{Giancola_2019_CVPR} is omitted for WOD comparison due to its costly training time.
As shown in Tab.~\ref{tab:nusc_waymo}, $M^2$-Track exceeds all the competitors under all categories, mostly by a large margin.
On such two challenging datasets with pervasive distractors and drastic appearance changes, the performance gap between previous approaches and $M^2$-Track becomes even 
larger (\eg more than \textbf{30\% }precision gain on Waymo Pedestrian).
Note that for large objects (\ie Truck, Trailer, and Bus), even if the predicted centers are far from the target (reflected from lower precision), the output BBoxes of the previous model may still overlap with the ground truth (results in higher success).
In contrast, the motion modeling helps to improve not only the success but also the precision by a large margin (\eg {\textbf{+23.43\%}} gain on Bus) for large objects.
Visualization results are provided in Fig.~\ref{fig:visualization} and the supplementary.

\subsection{Analysis Experiments}\label{sec:analysis}

\begin{table}[t]
  \centering
  \footnotesize
  \caption{Integration with Appearance Matching.}
  \vspace{-0.3cm}
    \begin{tabular}{c|cc}
    \toprule
    Method & Success & Precision \\
    \midrule\midrule
    PTT~\cite{shan2021ptt} & 67.80 & 81.80 \\
    V2B~\cite{hui20213d} & \textbf{70.50} & 81.30 \\
    \midrule
    $M^2$-Track & 65.49 & 80.81 \\
    $M^2$-Track + BAT~\cite{zheng2021box} & 69.22 \textcolor[rgb]{0.1,0.7,0.0}{\small $\uparrow$ 3.73} & 81.09 \textcolor[rgb]{0.1,0.7,0.0}{\small $\uparrow$ 0.28} \\
    $M^2$-Track + P2B~\cite{qi2020p2b} & 70.21 \textcolor[rgb]{0.1,0.7,0.0}{\small \textbf{$\uparrow$ 4.72}} & \textbf{81.80 }\textcolor[rgb]{0.1,0.7,0.0}{\small \textbf{$\uparrow$ 0.99}} \\
    \bottomrule
    \end{tabular}%
  \label{tab:appearanc_matching}%
  \vspace{-0.3cm}
\end{table}%
\begin{figure}[t]
  \centering
  \includegraphics[width=0.9\linewidth]{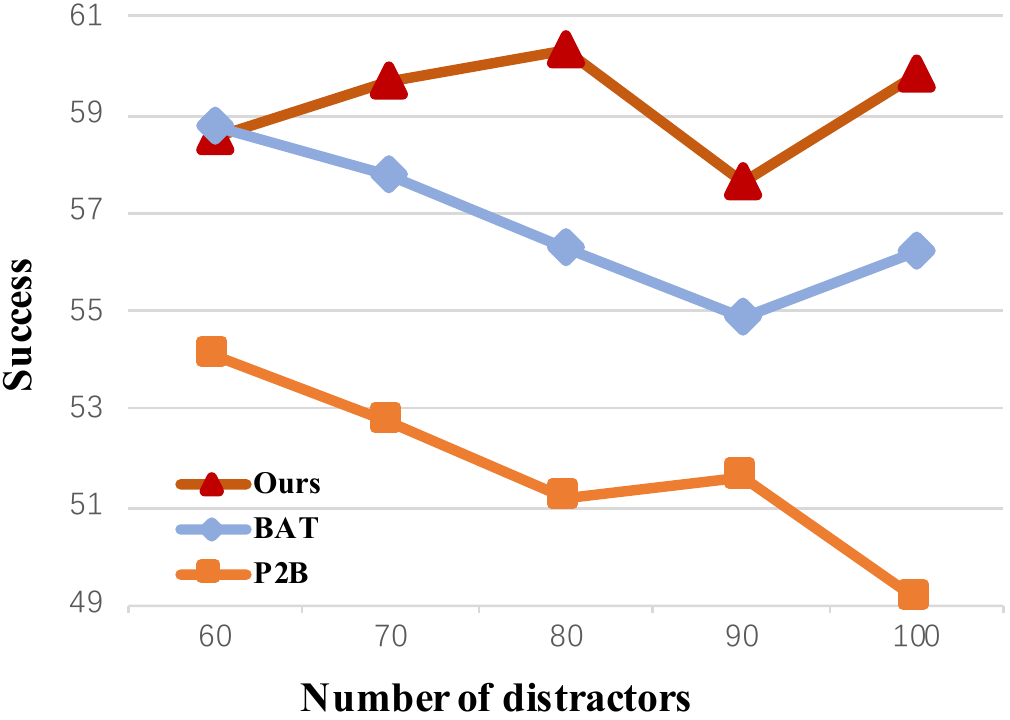}
  \caption{Robustness analysis with variant numbers of distractors.}
   \label{fig:distractors}
\end{figure}

In this section, we extensively analyze $M^2$-Track with a series of experiments.
Firstly, we compare the behaviors of $M^2$-Track and previous appearance-matching-based methods in different setups. 
Afterward, we equip $M^2$-Track with the previous appearance matching approaches to show its potential. 
Finally, we study the effectiveness of each component in $M^2$-Track.
All the experiments are conducted on the \textit{Car category of KITTI} unless otherwise stated.

\begin{table*}[htbp]
  \centering
  \footnotesize
  \caption{Results of $M^2$-Track when different modules are ablated. The last row denotes the full model. \textbf{Bold} denotes the largest change.}
  \vspace{-0.3cm}
    \begin{tabular}{cccc|cc|cc}
    \toprule
    {\multirow{2}{*}{\shortstack[c]{\\Box Aware\\ Enhancement}}} & 
    {\multirow{2}{*}{\shortstack[c]{\\Prev Box \\ Refinement}}} & 
    {\multirow{2}{*}{\shortstack[c]{\\Motion\\Classification}}} & 
    {\multirow{2}{*}{\shortstack[c]{\\Stage-II}}} & \multicolumn{2}{c|}{Kitti} & \multicolumn{2}{c}{NuScenes} \\
\cmidrule{5-8}          &       &       &       &  
{Success} & {Precision} & {Success} & {Precision} \\
    \midrule\midrule
               & \cmark     & \cmark     & \cmark     & 62.00 \textcolor[rgb]{0.7,0.0,0.0}{\small $\downarrow$ 3.49}   & 76.15 \textcolor[rgb]{0.7,0.0,0.0}{\small \textbf{$\downarrow$ 4.66}}&  53.68 \textcolor[rgb]{0.7,0.0,0.0}{\small \textbf{$\downarrow$ 2.17}}    &62.47 \textcolor[rgb]{0.7,0.0,0.0}{\small $\downarrow$ 2.62}  \\
    \cmark     &            & \cmark     & \cmark     & 64.23 \textcolor[rgb]{0.7,0.0,0.0}{\small $\downarrow$ 1.26}   & 78.12 \textcolor[rgb]{0.7,0.0,0.0}{\small $\downarrow$ 2.69}&    54.70 \textcolor[rgb]{0.7,0.0,0.0}{\small $\downarrow$ 1.15}   &  61.94 \textcolor[rgb]{0.7,0.0,0.0}{\small \textbf{$\downarrow$ 3.15}} \\
    \cmark     & \cmark     &            & \cmark     & 65.74 \textcolor[rgb]{0.1,0.7,0.0}{\small $\uparrow$ 0.25}     & 80.29 \textcolor[rgb]{0.7,0.0,0.0}{\small $\downarrow$ 0.52}&  54.88 \textcolor[rgb]{0.7,0.0,0.0}{\small $\downarrow$ 0.97}     & 64.40 \textcolor[rgb]{0.7,0.0,0.0}{\small $\downarrow$ 0.69} \\
    \cmark     & \cmark     & \cmark     &            & 61.29 \textcolor[rgb]{0.7,0.0,0.0}{\small \textbf{$\downarrow$ 4.20}}   & 77.31 \textcolor[rgb]{0.7,0.0,0.0}{\small $\downarrow$ 3.50}&    54.66 \textcolor[rgb]{0.7,0.0,0.0}{\small $\downarrow$ 1.99}   & 64.15 \textcolor[rgb]{0.7,0.0,0.0}{\small $\downarrow$ 0.94} \\
    \cmark     & \cmark     & \cmark     & \cmark     & 65.49                & 80.81 & 55.85 & 65.09 \\
    \bottomrule
    \end{tabular}%
  \label{tab:ablation}%
  \vspace{-0.3cm}
\end{table*}%

\noindent\textbf{Robustness to Distractors.}
Though achieving promising improvement on NuScenes and WOD, $M^2$-Track brings little improvement on the Car of KITTI.
To explain this, we look at the scenes of three datasets and find that the surroundings of most cars in KITTI are free of distractors, which are pervasive in NuScenes and WOD.
Although appearance-matching-based methods are sensitive to distractors, they provide more precise results than our motion-based approach in distractor-free scenarios. But as the number of distractors increases, these methods suffer from noticeable performance degradation due to ambiguities from the distractors.
To verify this hypothesis, we randomly add $K$ car instances to each scene of KITTI, and then re-train and evaluate different models using this synthesis dataset. 
As shown in Fig.~\ref{fig:distractors}, $M^2$-Track consistently outperforms the other two matching-based methods in scenes with more distractors, and the performance gap grows as $K$ increases.
Thanks to the box-awareness, BAT~\cite{zheng2021box} can aid such ambiguities to some extent.
But our performance is more stable than BAT's when more distractors are added.
Besides, the first row in Fig.~\ref{fig:visualization} shows that, when the number of points decreases due to occlusion, BAT is misled by a distractor and then tracks off course, while $M^2$-Track keeps holding tight to the ground truth.
All these observations demonstrate the robustness of our approach. 

\noindent\textbf{Influence of Motion Augmentation.}
We improve the performance of $M^2$-Track using the motion augmentation in training, which is not adopted in previous approaches.
For a fair comparison, we re-train BAT~\cite{zheng2021box} and P2B~\cite{qi2020p2b} using the same configurations in their open-source projects except additionally adding motion augmentation. 
Tab.~\ref{tab:augmentation} shows that motion augmentation instead has an adverse effect on both BAT and P2B.
Our model benefits from motion augmentation since it explicitly models target motion and is robust to distractors.
In contrast, motion augmentation may move a target closer to its potential distractors and thus harm those appearance-matching-based approaches.

\noindent\textbf{Combine with Appearance Matching.}
Although our motion-centric model outperforms previous methods from various aspects, appearance-matching-based approaches still show their advantage when dealing with distractor-free scenarios.
To combine the advantages of both motion-based and matching-based methods, we apply BAT/P2B as a ``re-tracker" to fine-tune the results of $M^2$-Track.
Specifically, we directly utilize BAT/P2B to search for the target in a small neighborhood of the $M^2$-Track's output.
Tab.~\ref{tab:appearanc_matching} confirms that $M^2$-Track can further benefit from appearance matching, even under this naive combination.
On KITTI Car, both combined models outperform the top-ranking PTT~\cite{shan2021ptt} by noticeable margins.
We believe that one can further boost 3D SOT by combining motion-based and matching-based paradigms with a more delicate design.

\noindent\textbf{Ablations.}
In Tab.~\ref{tab:ablation}, we conduct an exhaustive ablation study on both KITTI and NuScenes to understand the components of $M^2$-Track.
Specifically, we respectively ablate the \textit{box-aware feature enhancement}, \textit{previous BBox refinement}, \textit{binary motion classification} and $2^{nd}$ stage from $M^2$-Track.
In general, the effectiveness of the components varies across the datasets, but removing any one of them causes performance degradation. 
The only exception is the \textit{binary motion classification} used in the $1^{st}$ stage, which causes a slight drop on KITTI in terms of success.
We suppose this is due to the lack of static objects for KITTI's cars, which results in a biased classifier.
Besides, Tab.~\ref{tab:ablation} shows that $M^2$-Track keeps performing competitively even with module ablated, especially on NuScenes.
This reflects that the main improvement of $M^2$-Track is from the motion-centric paradigm instead of the specific pipeline design.


\section{More Discussion}\label[]{sec:discuss}
\noindent\textbf{Running Overheads.}
$M^2$-Track achieves exciting performance with just a simple PointNet~\cite{qi2017pointnet}.
Compared with other hierarchical backbones (\eg~\cite{qi2017pointnet++}) used in previous works, PointNet saves more computational overheads since it does not perform any sampling or grouping operations, which are not only time-consuming but also memory-intensive.
Therefore, $M^2$-Track runs \textbf{1.67$\times$} faster as the previous top-performer BAT~\cite{zheng2021box} (only consider model forwarding time) but saves \textbf{31.1\%} memory footprint.
Using a more advanced backbone (\eg \cite{qi2017pointnet++,thomas2019kpconv}) may further boost the performance but inevitably slows down the running speed.
Since we focus on online tracking, we prefer a simpler backbone to balance performance and efficiency.

\noindent\textbf{Limitations.}
Unlike appearance matching, our motion-centric model requires a good variety of motion in the training data to ensure its generalization on data sampled with different frequencies. For instance, our model suffers from considerable performance degradation if trained with 2Hz data but tested with 10Hz data because the motion distribution of the 2Hz and 10Hz data differs significantly. But fortunately, we can aid this using a well-design motion augmentation strategy.

\section{Conclusions}\label[]{sec:conclusion}
In this work, we revisit 3D SOT in LiDAR point clouds and propose to handle it with a new motion-centric paradigm, which is proven to be an excellent complement to the matching-based Siamese paradigm.
In addition to the new paradigm, we propose a specific motion-centric tracking pipeline $M^2$-Track, which significantly outperforms the state-of-the-arts from various aspects.
Extensive analysis confirms that the motion-centric model is robust to distractors and appearance changes and can directly benefit from previous matching-based trackers.
We believe that the motion-centric paradigm can serve as a primary principle to guide future architecture designs.
In the future, we may try to improve $M^2$-Track by considering more frames and integrating it with the appearance matching under a more delicate design.

\section*{Acknowledgment}
{\noindent This work was supported in part by NSFC-Youth  61902335,  by Key Area R\&D Program of Guangdong Province with grant No.2018B030338001, by the National Key R\&D Program of China with grant No.2018YFB1800800,  by Shenzhen Outstanding Talents Training Fund, by Guangdong Research  Project No.2017ZT07X152, by Guangdong Regional Joint Fund-Key Projects 2019B1515120039,  by the  NSFC 61931024\&81922046, by helixon biotechnology company Fund and CCF-Tencent Open Fund.}
{\small
\bibliographystyle{ieee_fullname}
\bibliography{ref}
}

\end{document}